%% file: iclr2026_conference.tex
\documentclass{article} 
\usepackage{iclr2026_conference,times}

\input{math_commands.tex}

\usepackage{hyperref}
\usepackage{url}
\usepackage{booktabs}
\usepackage{graphicx} 
\usepackage{multirow}
\usepackage{arydshln}    
\usepackage{float} 
\restylefloat{table}

\title{Probabilistic Modeling of Intentions in\\ Socially Intelligent LLM Agents}

\author{
\textbf{Feifan Xia}$^{1,2}$,
\textbf{Yuyang Fang}$^{3}$,
\textbf{Defang Li}$^{1}$,
\textbf{Yantong Xie}$^{4}$,
\textbf{Weikang Li}$^{5}$,
\textbf{Yang Li}$^{1*}$,\\
\textbf{Deguo Xia}$^{1}$,
\textbf{Jizhou Huang}$^{1}$\\[4pt]
$^{1}$Baidu Inc \quad
$^{2}$Imperial College London \quad
$^{3}$Zhejiang University \\
$^{4}$Carnegie Mellon University \quad
$^{5}$Peking University\\[3pt]
\texttt{\{xiafeifan, lidefang, liyang164, xiadeguo, huangjizhou01\}@baidu.com}\\
\texttt{feifan.xia23@imperial.ac.uk, fangyuyang@zju.edu.cn,}\\
\texttt{yantongx@andrew.cmu.edu, wavejkd@pku.edu.cn}
}

\iclrfinalcopy 

\begin{document}

\maketitle

\AtBeginEnvironment{quote}{\sloppy\hyphenpenalty=10000\exhyphenpenalty=10000}
\begin{abstract}
We present a probabilistic intent modeling framework for large language model (LLM) agents in multi-turn social dialogue. The framework maintains a belief distribution over a partner’s latent intentions, initialized from contextual priors and dynamically updated through likelihood estimation after each utterance. The evolving distribution provides additional contextual grounding for the policy, enabling adaptive dialogue strategies under uncertainty. Preliminary experiments in the SOTOPIA environment show consistent improvements: the proposed framework increases the \textit{Overall} score by \textbf{9.0\%} on \textit{SOTOPIA-All} and \textbf{4.1\%} on \textit{SOTOPIA-Hard} compared with the Qwen2.5-7B baseline, and slightly surpasses an oracle agent that directly observes partner intentions. These early results suggest that probabilistic intent modeling can contribute to the development of socially intelligent LLM agents.
\end{abstract}

\section{Introduction}

Social intelligence (Gweon et al., 2023; Mathur et al., 2024; Zhu et al., 2025) has emerged as a critical capability for large language models (LLMs) to adapt to complex social tasks. Whether in competitive settings like games, collaborative tasks like team projects, or mixed-motive scenarios like negotiation, success requires building predictive models of another agent’s behavior by inferring their hidden strategies and latent capabilities. Recent benchmarks assess these abilities from different perspectives, including 
\textit{MultiChallenge} (Zeng et al., 2024) for multi-turn reasoning and context following, 
\textit{MT-Bench} (Zheng et al., 2023) for automatic dialogue evaluation through model voting, 
and \textit{SimpleBench} (Liang et al., 2024) for commonsense, temporal, and social reasoning. Beyond general reasoning, the \textit{SOTOPIA} framework (Zhou et al., 2024) focuses specifically on social dialogue, evaluating agents along seven social dimensions such as goal completion, relationship maintenance, and norm compliance.

Existing approaches enhance LLMs’ social behavior primarily through policy optimization. Behavior cloning in \textit{SOTOPIA-pi} (Wang et al., 2024) and reinforcement learning in \textit{SOTOPIA-RL} (Yu et al., 2025) align model responses with high-scoring episodes in \textit{SOTOPIA-EVAL}. While effective for improving alignment, these methods treat dialogue generation as reward-driven imitation. In contrast, social dialogue is inherently an information-driven, multi-objective process: given limited and evolving information, an agent must optimize across several social objectives simultaneously. Even an oracle agent with full access to partner intentions remains static and cannot adaptively manage uncertainty.

We propose a probabilistic framework that represents dialogue context as a belief distribution over partner intentions and updates it iteratively through observed utterances. The evolving belief distribution serves as an auxiliary context that modulates the agent’s policy decisions. This design allows the agent to adapt its strategy as uncertainty changes, without requiring explicit parameter tuning or additional optimization.

\begin{figure}[H]
    \centering
    \includegraphics[width=0.9\linewidth]{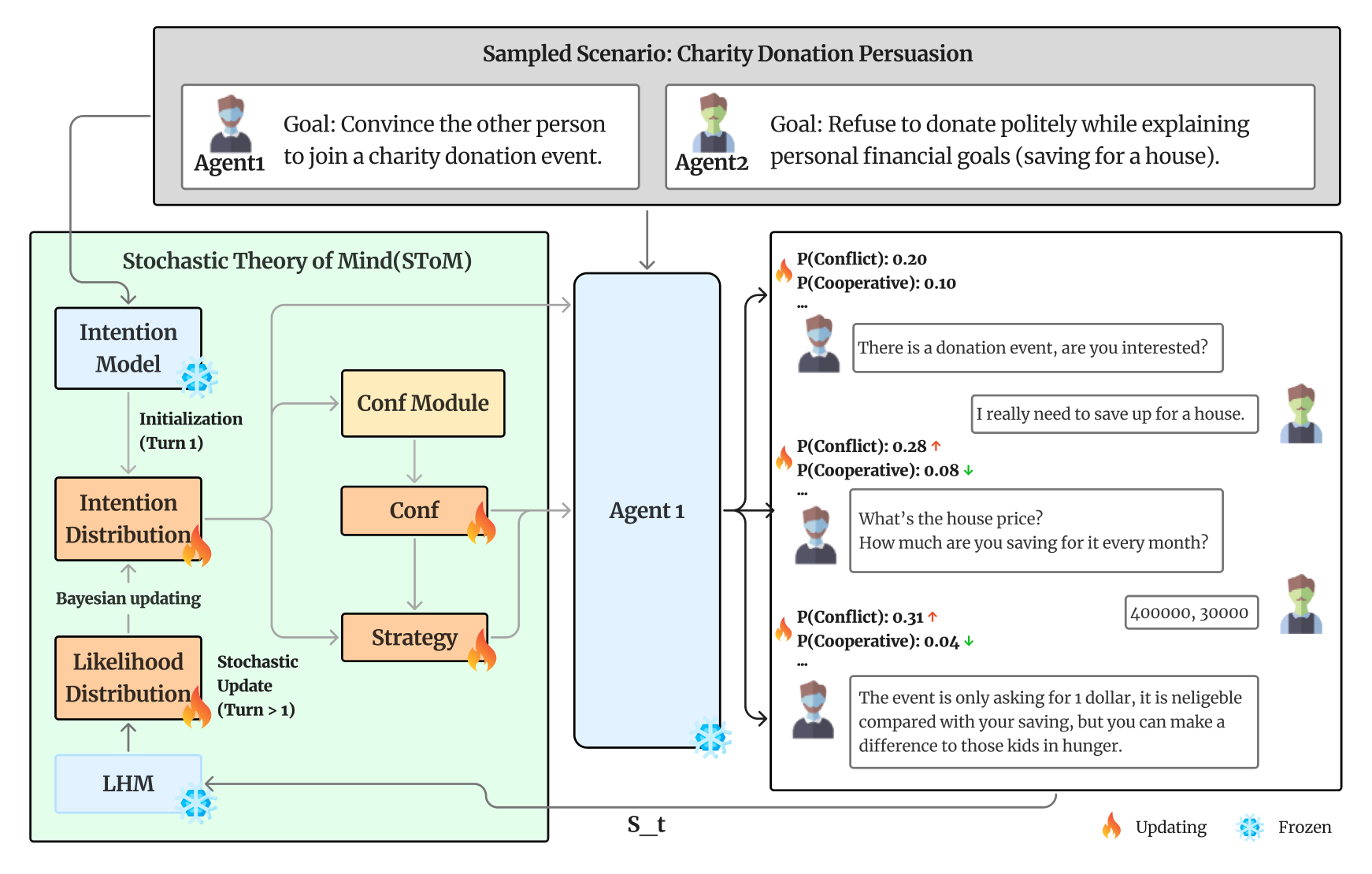}
    \caption{\textbf{Overview of the Stochastic Theory-of-Mind (SToM) framework.}
At each dialogue turn, the Intention Model (IM) produces a prior belief distribution over the partner’s possible intentions.
After observing the partner’s utterance, the Likelihood Model (LHM) updates the distribution via Bayesian inference.
The updated belief is serialized and appended to the policy model’s prompt, guiding subsequent actions.
This process enables belief-driven adaptation without any parameter training.
}
    \label{fig:placeholder}
\end{figure}

In preliminary evaluations on \textit{SOTOPIA-All} and \textit{SOTOPIA-Hard}, the framework improves the \textit{Overall} score by 9.0\% and 4.1\% respectively compared with the base model, and exceeds the oracle agent that receives ground-truth intentions.
\textbf{Our contributions are:}
(1) a probabilistic intent modeling formulation for LLM agents that explicitly represents and updates belief over partner intentions during dialogue;
(2) empirical evidence that maintaining such probabilistic intent representations improves multi-dimensional social performance even without additional training.

\section{Related Work}

\textbf{Social intelligence in LLM agents.} Recent work investigates improving the social capabilities of LLM agents through interaction and reinforcement learning. SOTOPIA-pi (Wang et al., 2024b) trains agents via self-interaction, motivated by human social learning (Tomasello, 2021; Gweon, 2021). SOTOPIA-RL (Yu et al., 2025) designs multi-dimensional rewards that capture utterance-level effects, extending earlier conversation-level RL rewards (Ndousse et al., 2021) and rule-based peer feedback (Pang et al., 2024). Preference-based methods such as SDPO (Kong et al., 2025) optimize alignment but ignore utterance-level dynamics. Applications include customer service (Pandya , Holia, 2023; Bamberger et al., 2023), education (Stamper et al., 2024; Nye et al., 2023), conflict resolution (Aggrawal , Magana, 2024), and team coordination (Li et al., 2023; Guo et al., 2024). These approaches emphasize skill acquisition and alignment but do not explicitly represent partner intentions.\\

\textbf{Theory of mind and intention modeling.} Another major line of research investigates how agents can infer and adapt to others’ hidden intention. Hypothetical Minds (Wong et al., 2023; Wang et al., 2024) introduces a ToM module that generates and evaluates hypotheses about other agents’ strategies, using LLMs for structured hypothesis search. While effective in grid-based games where latent features can be directly validated, this approach is less suited for open-ended dialogue where intentions must be inferred incrementally from soft linguistic evidence. A broader set of works explores opponent modeling in multi-agent reinforcement learning (Lowe et al., 2017; Yu et al., 2022a; Vezhnevets et al., 2020; Rabinowitz et al., 2018; Sclar et al., 2022; Huh, Mohapatra, 2024). These methods highlight the importance of intention modeling but often suffer from high sample complexity and limited generalization. In parallel, LLM-based agents have been deployed in embodied or multi-agent environments for long-horizon planning and coordination (Huang et al., 2022; Brohan et al., 2023a; 2023b; Rana et al., 2023; Liu et al., 2023a), collaborative social simulations (Park et al., 2023), subgoal planning (Li et al., 2023), and cooperative gameplay (Zhang et al., 2023a; 2023b). These approaches demonstrate the potential of LLMs for modeling others’ goals and behaviors, yet typically rely on heuristic prompting or task-specific structures rather than principled probabilistic frameworks.

\section{Method}

\subsection{Problem Formulation}
We study a 2-agent dialogue social interaction, modeled as a partially observable Markov decision process (POMDP), represented by the tuple $\langle S, A, O, T, Z, R \rangle$. 
Here, $S$ represents the set of possible social states, $A$ the action space, and $O$ the observation space. 
$T : S \times A \rightarrow S$ is the transition function. 
$Z : S \rightarrow O$ is the observation function. 
$R : S \times A \rightarrow \mathbb{R}$ is the reward function that reflects the overall quality of the social interaction with respect to each agent's private goal, such as successful persuasion or mutual understanding.

To incorporate Stochastic Theory of Mind (SToM), we extend this formulation with belief states. 
Let $\Theta$ represent the space of possible intentions/goals for the partner agent. 
We introduce a belief distribution $B_t(\theta)$ over the partner's latent intentions $\theta \in \Theta$ at time step $t$. 
Our augmented state space becomes $S' = S \times B$, where each state includes both the observable social context and the agent's belief distribution over the partner's intentions.

\subsection{Stochastic Theory of Mind (SToM) Framework}
Our SToM framework consists of three key components: 
\begin{enumerate}
    \item an \textbf{Intention Model (IM)} for generating and updating belief distributions,
    \item a \textbf{Likelihood Model (LHM)} for Bayesian belief updates, and
    \item a \textbf{confidence-aware action policy} that leverages uncertainty quantification.
\end{enumerate}

\subsubsection{Intention Model (IM)}
The Intention Model serves as the core belief reasoning component. 
Given the agent's observation history $h_t = \{o_1, o_2, \ldots, o_t\}$, the IM maintains a probability distribution over the partner's possible intentions:
\[
B_t(\theta) = P(\theta \mid h_t).
\]

\paragraph{Initialization Phase.}
At $t=0$, the IM analyzes the scenario context, agent backgrounds, and the focal agent's own goal to generate a set of plausible partner intentions 
$\Theta = \{\theta_1, \theta_2, \ldots, \theta_k\}$ (typically $k=3$--$5$). 
The initial belief distribution $B_0(\theta)$ is established based on scenario priors and complementary/adversarial goal reasoning.

\paragraph{Update Phase.}
For $t > 0$, beliefs are updated using Bayesian inference:
\[
B_{t+1}(\theta) \propto P(o_{t+1} \mid \theta) \cdot B_t(\theta),
\]
where $P(o_{t+1} \mid \theta)$ is provided by the Likelihood Model.

\subsubsection{Likelihood Model (LHM)}
The Likelihood Model estimates how probable a partner's observed action is given each hypothesized intention. 
For each intention $\theta_i \in \Theta$ and observed partner action $o_{t+1}$, the LHM computes:
\[
L_i = P(o_{t+1} \mid \theta_i).
\]

This likelihood captures the consistency between the partner's actual behavior and what would be expected under each intention hypothesis. 
The LHM leverages large language models to perform this reasoning, considering factors such as:
\begin{itemize}
    \item Semantic alignment between actions and intentions,
    \item Strategic consistency with goal-directed behavior,
    \item Contextual appropriateness given the dialogue history.
\end{itemize}

\subsubsection{Confidence-Aware Policy}
Our policy $\pi(a_t \mid s_t, B_t)$ explicitly considers belief uncertainty when selecting actions. 
We define confidence as:
\[
C_t = 1 - \frac{H(B_t)}{\log(|\Theta|)},
\]
where $H(B_t)$ is the Shannon entropy of the belief distribution, normalized by the maximum possible entropy.

\paragraph{High Confidence Regime ($C_t > \tau_{high}$).} 
When beliefs are concentrated, the agent pursues goal-directed actions based on the most likely partner intention.

\paragraph{Low Confidence Regime ($C_t < \tau_{low}$).} 
When beliefs are uncertain, the agent prioritizes information-gathering actions designed to elicit clarifying responses from the partner.

\paragraph{Medium Confidence Regime ($\tau_{low} \leq C_t \leq \tau_{high}$).} 
The agent balances goal pursuit with strategic information acquisition.

This confidence-aware approach naturally improves multiple evaluation dimensions:
\begin{itemize}
    \item \textbf{Knowledge Acquisition}: Low confidence triggers strategic questioning, improving information gathering.
    \item \textbf{Social Norms}: Uncertainty reduces impulsive actions that might violate social expectations.
    \item \textbf{Goal Achievement}: Higher confidence from better intention understanding leads to more effective goal-directed strategies.
\end{itemize}

\subsubsection{Implementation Details}
Our framework is implemented within the SOTOPIA environment. 
The IM and LHM are instantiated using GPT-4o, with carefully designed prompts that encourage:
\begin{enumerate}
    \item \textbf{Natural Questioning}: When confidence is low, prompts guide the agent to ask questions that feel conversational rather than interrogative.
    \item \textbf{Information Integration}: Prompts emphasize synthesizing new observations with existing beliefs.
    \item \textbf{Strategic Reasoning}: High-confidence scenarios encourage bold, goal-directed actions.
\end{enumerate}

The belief distribution is communicated to the acting agent through an enhanced ``Theory of Mind'' section that includes both intention hypotheses and their associated probabilities, enabling the agent to reason explicitly about uncertainty and plan accordingly.

\section{Experimental Setup}

\paragraph{Environment.}
We evaluate our framework in \textbf{SOTOPIA} (Zhou et al., 2024), an open-ended environment for goal-oriented social dialogue.
The benchmark contains 90 diverse scenarios (\textit{SOTOPIA-All}) and a subset of 14 more challenging ones (\textit{SOTOPIA-Hard}).
Each episode involves two agents with private intentions, and performance is assessed by the \textbf{SOTOPIA-EVAL} metrics:
\textit{Goal}, \textit{Believability}, \textit{Relationship}, \textit{Knowledge}, \textit{Social Norms}, \textit{Secret}, and \textit{Financial Benefit}.
All results are reported as GPT-4o-based automated evaluations, averaged over multiple seeds with $p<0.05$ significance.

\paragraph{Models.}
We compare three configurations using the same LLM backbone ({Qwen 2.5-7B-Instruct) as the dialogue policy.
\begin{itemize}
    \item \textbf{Base:} Qwen 2.5-7B-Instruct without additional modules. 
    \item \textbf{SToM (Ours):} adds a probabilistic intent modeling module that maintains a belief distribution over partner intentions and updates it online via a likelihood model (LHM). The distribution is serialized and injected into the policy prompt at each turn. Both IM and LHM operate without training or gradient updates.  
    \item \textbf{Oracle Agent (Upper Bound):} an oracle agent conditioned with the ground-truth intention of its partner at every turn.This configuration represents a privileged-information upper bound for comparison.  
\end{itemize}
All agents interact with GPT-4o partners under identical sampling and temperature settings.

\paragraph{Evaluation Protocol.}
Each configuration is evaluated on the full 90-episode \textit{SOTOPIA-All} and 14-episode \textit{SOTOPIA-Hard} sets.
The \textit{Overall} score is computed as the mean of the seven normalized SOTOPIA-EVAL dimensions.
We report relative percentage improvements with respect to the Base model.

\section{Results}

\begin{table}[H]
\centering
\caption{\textbf{Main results.} The highest score is highlighted in bold (excluding API models). 
The reported results are averaged over 90 runs for \textit{SOTOPIA-All} and 14 runs for \textit{SOTOPIA-Hard} 
(both statistically significant with $p<0.05$).}
\vspace{2mm}
\begin{tabular}{lcccccc}
\toprule
 & \multicolumn{5}{c}{\textit{GPT-4o-as-Partner}} \\
\cmidrule(lr){2-6}
\multirow{2}{*}{\textbf{Models}} & 
\multicolumn{2}{c}{\textbf{SOTOPIA-All}} & & 
\multicolumn{2}{c}{\textbf{SOTOPIA-Hard}} \\
\cmidrule(lr){2-3} \cmidrule(lr){5-6}
 & Goal $\uparrow$ & Overall $\uparrow$ & & Goal $\uparrow$ & Overall $\uparrow$ \\
\midrule

GPT-4o & 8.19 & 3.76 & & 6.97 & 3.46 \\
Claude-Sonnet-3.5 & 8.42 & 3.77 & & 6.64 & 3.30 \\
Deepseek-V3-671B & 8.14 & 3.72 & & 6.69 & 3.31 \\
\hdashline
OpenAI-o1 & 8.09 & 3.69 & & 6.65 & 3.20 \\
OpenAI-o3-mini & 7.96 & 3.61 & & 6.33 & 2.98 \\
DeepSeek-R1 & 7.92 & 3.49 & & 6.20 & 2.95 \\
QwQ-32B & 7.80 & 3.47 & & 6.19 & 2.91 \\
Gemini-2.0-flash-thinking & 7.82 & 3.56 & & 6.81 & 3.27 \\
\hdashline
\textbf{Qwen2.5-7B-it} & 6.71 & 3.13 & & 5.90 & 2.90 \\
\quad w/ SToM(ours) & 7.32 & \textbf{3.45} & & 5.93 & \textbf{3.14} \\
\quad w/ Ground truth & \textbf{7.62} & 3.38 & & \textbf{6.14} & 3.03 \\
\bottomrule
\end{tabular}
\end{table}

\paragraph{Quantitative results.}
Table 1 summarizes the averaged scores on both splits.
Without any parameter training, \textbf{SToM} achieves consistent improvements across all metrics.
Compared with the Baseline (Qwen2.5-7B-it), SToM increases the \textit{Overall} score by \textbf{+9.0\%} on \textit{SOTOPIA-All} and \textbf{+4.1\%} on \textit{SOTOPIA-Hard}.
Notably, SToM also surpasses the \textit{Oracle Agent}, which directly accesses the ground-truth of opponent's intentions, by \textbf{+0.6\%} and \textbf{+1.7\%} respectively.
This result indicates that stochastic belief updating can yield advantages beyond deterministic oracle conditioning.

\paragraph{Analysis.}
The oracle agent, while possessing perfect knowledge of partner intentions, follows a fixed interaction trajectory and lacks adaptability to uncertainty.
In contrast, SToM’s stochastic belief updates encourage exploratory strategies that dynamically balance competing social objectives (e.g., persuasiveness vs.\ relationship maintenance).
This leads to more information-efficient dialogue and higher aggregate social performance.
The gain on \textit{SOTOPIA-Hard} suggests improved robustness in scenarios with ambiguous or conflicting goals.

\paragraph{Summary.}
These results demonstrate that explicit probabilistic intention maintenance can enhance the social competence of LLM agents even without training or external rewards.
The observed improvements validate the proposed stochastic theory-of-mind (SToM) framework as a lightweight yet effective agentic framework.

\section{References}

Aggrawal, P., and Magana, A.  
Dialogue Agents for Conflict Resolution.  
Computers in Human Behavior, 2024.

Bamberger, T., et al.  
Evaluating Social Competence in Conversational Agents.  
ACM Transactions on Interactive Intelligent Systems, 2023.

Brohan, A., et al.  
RT-X: Scaling Robotics Foundation Models with Real-World Data.  
arXiv preprint arXiv:2306.12973, 2023.

Gweon, H.  
Cognitive Foundations of Social Learning in Children and Machines.  
Trends in Cognitive Sciences, 2021.

Gweon, H., et al.  
The Computational Foundations of Social Intelligence.  
Trends in Cognitive Sciences, 2023.

Guo, R., et al.  
Collaborative Language Agents for Multi-Agent Communication.  
International Conference on Learning Representations (ICLR), 2024.

Huang, W., Brohan, A., et al.  
Decision Transformer: LLMs for High-Level Planning and Control in Robotics.  
IEEE International Conference on Robotics and Automation (ICRA), 2022.

Huh, M., and Mohapatra, S.  
Theory of Mind Modeling for Adaptive Multi-Agent Systems.  
International Conference on Learning Representations (ICLR), 2024.

Kong, J., et al.  
SDPO: Preference-Based Social Alignment of Dialogue Agents.  
arXiv preprint arXiv:2503.01942, 2025.

Li, C., et al.  
Multi-Agent Team Coordination in Open-Ended Dialogue Tasks.  
arXiv preprint arXiv:2309.09780, 2023.

Li, H., et al.  
SAMA: Subgoal-Aware Multi-Agent Coordination via LLM Planning.  
arXiv preprint arXiv:2310.10201, 2023.

Liang, S., et al.  
SimpleBench: Benchmarking Commonsense, Temporal, and Social Reasoning in LLMs.  
arXiv preprint arXiv:2408.01491, 2024.

Liu, Y., et al.  
Embodied Multi-Agent Environments for Large Language Models.  
arXiv preprint arXiv:2307.04750, 2023.

Lowe, R., Wu, Y., Tamar, A., Harb, J., Abbeel, P., and Mordatch, I.  
Multi-Agent Actor-Critic for Mixed Cooperative-Competitive Environments.  
Conference on Neural Information Processing Systems (NeurIPS), 2017.

Mathur, N., et al.  
Measuring and Modeling Social Intelligence in AI Systems.  
Nature Machine Intelligence, 2024.

Ndousse, K., et al.  
Learning Socially Aligned Reinforcement Learning Policies.  
Conference on Neural Information Processing Systems (NeurIPS), 2021.

Nye, B., et al.  
LLM Tutors for Personalized Education.  
arXiv preprint arXiv:2310.05641, 2023.

Pang, L., et al.  
Stable Alignment via Rule-Based Peer Feedback.  
arXiv preprint arXiv:2404.10811, 2024.

Pandya, R., and Holia, D.  
Conversational AI for Customer Service Applications.  
IEEE Transactions on Affective Computing, 2023.

Park, J., et al.  
Generative Agents: Interactive Simulacra of Human Behavior.  
arXiv preprint arXiv:2304.03442, 2023.

Rabinowitz, N. C., Perbet, F., Song, H. F., Zhang, C., Eslami, S. M. A., and Botvinick, M.  
Machine Theory of Mind.  
International Conference on Learning Representations (ICLR), 2018.

Rana, R., et al.  
LLMs for Robotic Planning under Partial Observability.  
International Conference on Learning Representations (ICLR), 2023.

Sclar, M., et al.  
Toward Theory-of-Mind Modeling in Multi-Agent Reinforcement Learning.  
Association for the Advancement of Artificial Intelligence Conference (AAAI), 2022.

Stamper, J., Nye, B., et al.  
Dialogue Systems for Educational Tutoring.  
International Conference on Artificial Intelligence in Education (AIED), 2024.

Tomasello, M.  
The Adaptive Origins of Human Social Learning.  
Annual Review of Psychology, 2021.

Vezhnevets, A. S., et al.  
OPRE: Observationally Predictive Representation Learning for Multi-Agent Environments.  
arXiv preprint arXiv:2006.12903, 2020.

Wang, H., Gweon, H., and Yu, Z.  
SOTOPIA-pi: Interactive Learning of Socially Intelligent Language Agents.  
arXiv preprint arXiv:2405.09978, 2024.

Wong, A., Wang, Z., and Park, J.  
Hypothetical Minds: Theory-of-Mind Inference in Multi-Agent Environments.  
arXiv preprint arXiv:2311.09764, 2023.

Yu, C., Xu, H., and Wang, Z.  
MAPPO: Multi-Agent Proximal Policy Optimization.  
Conference on Neural Information Processing Systems (NeurIPS), 2022.

Yu, Z., Zhou, Y., Wang, H., and Gweon, H.  
SOTOPIA-RL: Reward Design for Social Intelligence in Large Language Models.  
arXiv preprint arXiv:2501.05892, 2025.

Zeng, X., et al.  
MultiChallenge: Benchmarking Multi-Turn Reasoning and Instruction Following.  
arXiv preprint arXiv:2405.04882, 2024.

Zhang, R., et al.  
ProAgent: Zero-Shot Multi-Agent Coordination via LLM Intent Inference.  
arXiv preprint arXiv:2311.05622, 2023.

Zhou, Y., Wang, H., Gweon, H., and Yu, Z.  
SOTOPIA: Interactive Evaluation for Social Intelligence in Language Agents.  
arXiv preprint arXiv:2403.15347, 2024.

Zheng, L., Chiang, W.-L., et al.  
MT-Bench: Multi-Turn Benchmark for Chatbot Evaluation.  
arXiv preprint arXiv:2306.05685, 2023.

Zhu, X., et al.  
Evaluating Social Reasoning in Large Language Models.  
arXiv preprint arXiv:2504.06721, 2025.

\end{document}

%% file: math_commands.tex
\usepackage{amsmath,amsfonts,bm}

\def\eqref#1{equation~\ref{#1}}

\def\1{\bm{1}}

\DeclareMathAlphabet{\mathsfit}{\encodingdefault}{\sfdefault}{m}{sl}
\SetMathAlphabet{\mathsfit}{bold}{\encodingdefault}{\sfdefault}{bx}{n}

